# Geocoding Multilingual Texts: Recognition, Disambiguation and Visualisation

**Bruno Pouliquen, Marco Kimler, Ralf Steinberger, Camelia Ignat, Tamara Oellinger, Ken Blackler, Flavio Fluart, Wajdi Zaghouani, Anna Widiger, Ann-Charlotte Forslund, Clive Best**

European Commission - Joint Research Centre
Via Enrico Fermi, 1
21020 Ispra (VA), Italy
Email: Firstname.Lastname@jrc.it; URL: http://www.jrc.it/langtech; http://press.jrc.it/NewsExplorer

**Abstract**
We are presenting a method to recognise geographical references in free text. Our tool must work on various languages with a minimum of language-dependent resources, except a gazetteer. The main difficulty is to disambiguate these place names by distinguishing places from persons and by selecting the most likely place out of a list of homographic place names world-wide. The system uses a number of language-independent clues and heuristics to disambiguate place name homographs. The final aim is to index texts with the countries and cities they mention and to automatically visualise this information on geographical maps using various tools.

## 1. Introduction

Geocoding texts is one aspect of Named Entity Recognition, which is a known research area (see MUC conferences). Our method differs from previous work in that the tool currently recognises geographical information in 16 languages (ar, cs, da, de, en, es, et, fi, fr, it, nl, pt, ro, ru, sl, sv)[1]. Additionally, instead of simply stating that a certain string is a place, we aim at disambiguating homographic place names. The current tool uses new disambiguation procedures (Kimler, 2004) compared to our previous exploratory work (Pouliquen et al, 2004). It is part of an operational application called NewsExplorer (http://press.jrc.it/NewsExplorer) which currently automatically builds up knowledge from newspaper articles in 13 languages (soon 15). In the course of daily news analysis (analysing 15,000 articles per day), the NewsExplorer system identifies places and other named entities. This information is then analysed to calculate relationships between entities, e.g. the most important people mentioned in the context of a certain country. Moreover, the geographical information allows to display information on a geographical map. NewsExplorer currently uses three different visualisation techniques that are explained in this paper.

## 2. Place Name Recognition

The place name recognition software is part of a larger text analysis and Named Entity Recognition software toolset, where person and organisation names are identified first and place names are identified only if they have not already been identified as other name types. This helps to avoid wrong hits in strings like 'George Bush' or 'Kofi Annan', where we do not want to identify the strings 'Bush' (city of Louisiana) and 'Annan' (small city in Scotland) as place names.

In our system, place names are identified exclusively through gazetteer lookup procedures and subsequent disambiguation or elimination. Contextual patterns are not used because, unlike for person names, geographical references have no or extremely weak contextual clues (Mikheev et al. 1999). For European languages, only uppercase words are looked up and matched against the database of names, while for languages like Arabic or Farsi, which do not distinguish case, every word needs to be looked up in the database to see whether it is a potential place name. An additional step allows us to use specific rules to handle declensions of place names (e.g. *Parisului* in Romanian means 'of Paris' where only 'Paris' is in our gazetteer). Those rules can even handle some frequent inhabitant names (e.g. *Les Parisiens*), or even more complicated rules like the Arabic string الباريســــون [alba-Riziu:n] (the Paris inhabitants), which is automatically transformed into بـــاريس [baRi:z] (Paris). These rules are currently implemented as a set of regular expression substitutions.

As our system relies mainly on the content of the gazetteer, we combined the content of three sources to build our own database:

Global Discovery database of place names (Global Discovery 2006), which contains over half a Million place names world-wide, in both English and the local language (but only in Roman script). It also contains latitude, longitude and size-category information.

The multilingual KNAB database (KNAB 2006), which contributed with large numbers of exonyms (place name in languages other than the local ones – Venice/Venise/Venedig), historical (Constantinople/Istanbul) and linguistic (Istamboul/Istanbul/İstanbul) variants.

A European commission internal document that lists, for each country, the name of the capital and of the inhabitant names, the currency, and the country adjective in the previous 11 official languages (e.g. France/French/Paris, Francia/Francese/Parigi)[2].

## 3. Disambiguation / Filtering

Geocoding is not only a lookup of every single word of a text in a gazetteer. The result of the place name lookup (*geoparsing*) then needs to be disambiguated.

The full disambiguation process uses (1) the fact that the place name is part of a known person name, (2) the importance of the place, (3) the main countries the text is about, (4) the presence or not in a geo-stop-word list, (5) the words in the immediate lexical context, (6) the minimum kilometric distance to any other place.

---

[1] Most of the Commercial systems, like MetaCarta (http://www.metacarta.com/), are monolingual.

[2] This document is partially available at http://publications.eu.int/code/en/en-370101.htm

### 3.1 Person Name Ambiguity

Table 1 lists known persons where the last name and the first name are also cities somewhere in the world. 'Javier' being a city in Spain and 'Solana' a city in the Philippines, a place name lookup in a sentence containing 'EU-general secretary Javier Solana' would recognise these two cities.

The solution we have undertaken consists of recognising first the person names in the text, and then of ignoring potential place names which are either a first or a last name in the text. We end up with a binary filter: all place names being part of a person name are left out.

| Name | City: Country |
|---|---|
| George Bush | *Georges:* S. Africa |
| | *Bush:* USA |
| Tony Blair | *Tony:* USA |
| | *Blair:* Malawi |
| Kofi Annan | *Kofi:* Mali |
| | *Annan*: Scotland |
| Javier Solana | *Javier:* Spain |
| | *Solana*: Philippines |
| Lance Armstrong | *Lance:* Mozambique |
| | *Amstrong*: Argentina |
| James Joyce | *James:* Zambia |
| | *Joyce:* USA |
| Jack Nicholson | *Jack*: South Africa |
| | *Nicholson:* Australia |
| Henry Ford | *Henry*: Haiti |
| | *Ford*: Ireland |

Table 1: Famous persons and their corresponding homographic place names.

### 3.2 Importance of the Place

When an ambiguity exists between two places (Paris-France and Paris-Texas for example), the most important place will have the higher weight. As a result: Paris-France will always be chosen over Paris-Texas unless some other clues trigger Paris-Texas.

The information we get from Global Discovery associates a "class" attribute from 1 to 6 to each entry of our database, we added the "class 0" to country names.

| Class | Explanation | Example | Weight |
|---|---|---|---|
| 0 | country name | Italy | 80 |
| 1 | capital | Rome | 80 |
| 2 | main city | Milan | 80 |
| 3 | province level | Varese | 30 |
| 4 | small city | Sesto Calende | 20 |
| 5 | village | Ispra | 10 |
| 6 & more | small settlement, hamlet | - | 5 |

Table 2: Class attribute associated with each entry in our database.

| Place name | Number of cities with this name |
|---|---|
| Aleksandrovka | 244 |
| San Antonio | 205 |
| Santa Rosa | 199 |
| ... | |
| San Francisco | 102 |
| Buenos Aires | 88 |
| Washington | 32 |
| London | 18 |
| Berlin | 15 |
| Paris | 15 |
| Rome | 15 |
| Moscow | 12 |

Table 3: Some frequent place names and their number of occurrences in the Global Discovery Gazetteer.

### 3.3 Main Countries the Text is about (Country Context)

Two or more different places could share the same name. Table 3 gives an overview of highly ambiguous place names. Places with the same name can sometimes only be disambiguated using the context of the text. The most important aspect of disambiguation consists of looking at the countries the text is talking about, before trying to recognise small cities. Currently we are using two techniques to set the context of the text: setting it to the publishing place and/or using *shallow-geoparsing*.

When analysing newspaper text, the country of origin of the article is often known so we can set this context (a newspaper article from a Zimbabwean newspaper is more likely to speak about Zimbabwe).

The *shallow-geoparsing* method consists of extracting very important unambiguous place names from the text. This first pass will return the countries the text is about (the context of the text). In order to build this tool we have selected the country names (class 0 entries) and the main cities (classes 1 and 2 in Table 2), where we ignore possible ambiguous names (e.g. 'Monaco' in Italian language designing both Munich, the German city, and the small country).

A second pass, called *deep-geoparsing*, will then recognise smaller cities only if they are located in the countries mentioned in the context (country of origin of the text or the result of the *shallow-geoparsing*).

### 3.4 Geo-Stop Words

Many place names are ambiguous in the sense that they are homonymic with other words in natural language, Table 4 lists some obvious examples of city names that are also among the 30 most frequent words in English, French and German.

With big gazetteers like ours, it is quite likely to encounter many common words as place names. Moreover a gazetteer is supposed to be extended and updated regularly. Therefore we built a geo-stop word detector tool. The tool analyses lists of known first names and proposes to tag those which could also be place names as stop words. The other functionality consists of compiling a reference list of words occurring in lowercase in a big corpus and then categorise as stop words the place names appearing in the list with a frequency above a given threshold. Currently, our stop word list contains 5489 place names.

| English | | French | | German | |
|---|---|---|---|---|---|
| *Place name* | *Country* | *Place name* | *Country* | *Place name* | *Country* |
| And | Iran | De | Burkina Faso | Die | France |
| To | Ghana | Du | Ghana | Den | Ethiopia |
| Be | India | Un | Russia | Zu | Zaire |
| By | Sweden | Une | Colombia | Ist | Hungary |
| Are | Nigeria | Est | Netherlands | Im | Russia |
| This | France | Il | Iran | Dem | Cameroon |
| But | Afghanistan | Au | Austria | Als | Denmark |
| Had | Oman | Par | U.K. | Auch | France |
| She | India | Sur | Oman | An | Mexico |
| We | Zaire | Pas | Turkey | Aus | Namibia |

Table 4: Some examples of place names homographic to very frequent words.

### 3.5 Words in the Immediate Lexical Context

Many geographical references, especially the ones representing smaller places like villages, have often a word indicating the place type in their surroundings. Such words are 'city', 'village', 'town' etc. We have experimented with the usage of such trigger words to give more weight to ambiguous place names. However, the input of this disambiguation method did not really improve the results. An explanation could be that the lexical context of place names is more in the part-of-speech information than really in the surrounding words ("And is beautiful" could mean the city of And – Iran – is beautiful, while "She is beautiful" does not mean that the city of She – India – is beautiful).

A first evaluation we launched did not yield better results, so that we decided not to implement this rule (knowing also that a proper implementation of it would need additional language-dependent rules).

### 3.6 Minimum Kilometric Distance

The "context" of the text described above relies mainly on the country the place belongs to. We are planning to extend it to the region/district it belongs to also, but this is sometimes not enough. Not only two villages of the same administrative unit could share a name (two villages called "Saint Cyr" in France – Saint-Cyr-la-rivière and Saint-Cyr-sous-Dourdan – are in the same department), but the kilometric distance could also be a good indicator for further disambiguation. An example is "from Warsaw to Brest" where *Brest* triggers equally the cities in France and in Belarus, but the *Brest* in Belarus is only 200 kilometres from Warsaw, while the French port is about 2000 kilometres away.

The tool computes the minimum kilometric distance between ambiguous places and unambiguous one. The distance is computed using latitude/longitude information, according to the formula by Sinnott (1984).

### 3.7 Combination of the Various Disambiguation Heuristics

The optimal parameters for the combinations of the 5 heuristics were manually set in order to maximise the performance on a development set of texts (different from the test set described in the next section). We have set two filtering heuristics and 3 disambiguation heuristics.

*Filtering heuristics:*
The geo-stop word list and person name ambiguity are currently set as filters. If a place name belongs to a person name or to the geo-stop list, it will never be recognised.

*Disambiguation heuristics:*
When a place name can trigger different places around the world, the program has to take a decision. The solution consists of computing a score for each alternative to this ambiguous place, so that the place having the highest score will be chosen.

For computing the score, the currents settings are:
Score = *classScore* [80,30,20,10,5]
+ 100 (if country in context)
+ 20 · *kilometricWeight*()

where:
*classScore* is a given score depending on the importance of the place, which is coded by the *class* attribute (as mentioned in Table 2): 80 for country name, capital or big city, 30 for province level city, 20 for small cities, 10 for villages, 5 for settlements;
*kilometricWeight* is the minimum distance between the place and all unambiguous places. This distance is weighted using the arc-cotangent formula, as defined by Bronstein et al. (1999), with an inflexion point set to 300 kilometres[3], as shown in Equation 1.

$$kmweight(d) = \frac{1}{arcCot(-\frac{300}{100})} arcCot(\frac{d-300}{100})$$

Equation 1: Kilometric distance weight: given a kilometric distance d between two places, return a weight between 0 and 1 indicating the "closeness" of two places.

## 4. Evaluation

### 4.1 Test Set

To our knowledge, no corpus currently exists to evaluate place name disambiguation. We must mention the nice initiative of Leidner (2004) who intends to build such a resource, although only for English. As no resource is yet available to evaluate the functionalities of our system, we decided to build our own multilingual evaluation set: we have chosen 162 newspaper stories out of the Europe Media Monitor (Best et al. 2002) containing possible small or ambiguous places. The advantage of such a corpus is that the articles were taken from comparable stories in German, English, Spanish, French and Italian. For information, the stories are listed in Table 5.

For comparing the results, we use the known metrics:

---

[3] Empirical experiments showed that distances of less than 200 km are very significant, and distances more than 500 km do not make a big difference. Therefore, we have chosen the inflexion point between those two values: 300.

*Precision = (number of relevant documents retrieved) / (number of documents retrieved)*

*Recall = (number of relevant documents retrieved) / (number of relevant documents)*

*F-measure = 2 (Precision · Recall) / (Precision + Recall).*

| Story | Languages | | | | | |
|---|---|---|---|---|---|---|
| | de | en | es | fr | it | all |
| Arrest of Islamic militants believed to have been planning a bomb attack during a NATO summit in Istanbul (3/05/2004) | 10 | 10 | 7 | 5 | 3 | 35 |
| Chechen President Kadyrov killed in a bomb explosion in Grozny (9/05/2004). | 10 | 10 | 10 | 10 | 10 | 50 |
| Earthquake in Northern Iran (29/05/2004) | 10 | 10 | 5 | 4 | 5 | 34 |
| Former US-president Ronald Reagan died (6/06/2004). | 10 | 10 | 8 | 6 | 8 | 42 |

Table 5: Test set composition

### 4.2 Evaluation of Disambiguation Contribution

We launched a first evaluation on the effect of our disambiguation techniques. Using the same gazetteer, we launched two experiments on the test set, the first one just looking up all place names, the second using our disambiguation techniques (except the *words in the immediate lexical context* as we do not use it any more). The first experiment yielded an F-measure of 55.4%, the second an F-measure of 77%, which clearly proves that disambiguation is compulsory when dealing with such a number of places (see Table 6).

We must mention that we evaluated the heuristics in five languages (English, German, Italian, French and Spanish). They worked similarly well in all languages, as shown in Table 7.

| Disambiguation technique | F-measure |
|---|---|
| None | 55.4% |
| Geo-context (country in the text) | 66.7% |
| Importance of the place | 75.9% |
| Minimum kilometric distance | 68.7% |
| Filtering by person names | 64.2% |
| Geo Stop list | 61.1% |
| All techniques | 77 % |

Table 6: F-measure achieved using each individual disambiguation process.

### 4.3 Evaluation of the Geocoding

The results were on average 77% precision, 77% recall (77% F-measure). See Table 7 for details about performance depending on the language. We ran the same evaluation with our previous tool (Pouliquen et al, 2004) where the F-measure was only 38% due to the complexity and ambiguity of the test set. This confirms the contribution of our disambiguation techniques.

| Language | Precision | Recall | F-measure |
|---|---|---|---|
| *German* | 80 | 68 | 73 |
| *English* | 91 | 78 | 84 |
| *Spanish* | 75 | 80 | 76 |
| *French* | 68 | 74 | 70 |
| *Italian* | 70 | 85 | 76 |
| *Average* | 77 | 77 | 77 |

Table 7: Geocoding evaluation results across languages.

We must highlight that the test set used for this evaluation contains a lot of ambiguous place names (the set was built to test disambiguation). Our previous algorithm, on a more generic test set (a random selection of 48 English articles), had better results: precision, recall, F-measure respectively equal to 98%, 90%, 94%, see (Pouliquen et al., 2005).

## 5. Visualisation of the Automatically Identified Place Names

As homographic places have been disambiguated in the previous step and latitude-longitude information is available, geographical references found in a text or in a cluster of related texts can be visualised using maps. We have experimented with various visualisation methods. Three of them are currently in use: flash animation-based world map, GoogleEarth interface and interactive SVG maps.

### 5.1 WorldKit

The NewsExplorer system displays a world map (Figure 1) produced using the freely available tool WorldKit (2006). WorldKit uses as input an RSS file (Xml format syndicating a list of articles) where each of the articles contains a unique latitude/longitude indicating the place where the event has taken place.

The resulting map is displayed using a Macromedia Flash animation embedded in the HTML page of the NewsExplorer.

### 5.2 GoogleEarth

Additionally, the user can browse the events of the day using the GoogleEarth interface (Figure 2). By using GoogleEarth we create a straightforward method of comparing and relating news and other information available (see GoogleEarth 2006). This additional data includes geographical data such as satellite imagery, borders, populated places and their sizes, infrastructure, etc. Many other dynamic sources are also available, including earthquakes, disease outbreak monitoring, World Heritage sites, etc.

We convert NewsExplorer daily data as specified in the KML 2.0 specification (KML 2006). For each news story a KML folder with a title and description is created. If the story includes geographical entities, then a list of KML *placemarks* is created underneath. Each description has a link to NewsExplorer, so further information can be shown. Three different icons are used to represent placemarks. They show the importance (media coverage) of a story: more than 20 , between 10 and 20  and up to 10  articles.

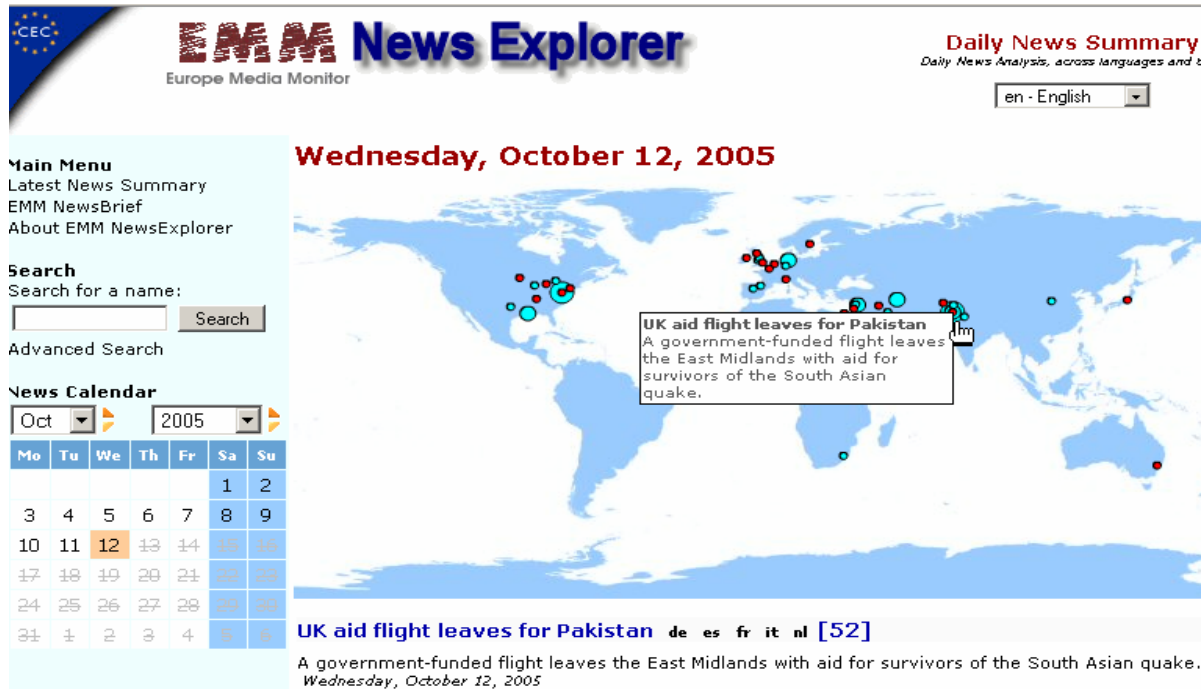

Figure 1: Visualisation using Worldkit, as used on the publicly accessible NewsExplorer system (http://press.jrc.it/NewsExplorer).

In order to display more precise information, we have chosen to leave out the small places (classes 4 to 6).

### 5.3 Interactive SVG Maps

Finally, a JRC-internal, interactive map uses the Support Vectors Graphics format (Eisenberg 2002) to highlight places on a map and to let users access the textual information interactively (Figure 3). Additionally, the highlighted countries and dots of varying sizes for cities found, plus people's names and subject domain-specific terminology found in the text can be listed on the map (Ignat et al., 2005).

### 6. Conclusion

The tool is now used on a daily basis in the NewsExplorer application. We have proved that it can easily be adapted to new languages (Arabic for example). Even if the geo-coding performance could still be improved, the evaluation has shown that we already have achieved acceptable results compared to the complexity of the task.

The various geographical maps are well perceived by users as classifying news by location is rather intuitive. It helps users to get a quick overview of what is happening in the world. Additionally, the visualisation also helped us to iteratively improve our recognition methods.

In addition to the visualisation on geographical maps, the geocoding of news adds useful meta-information to each article, which then allows NewsExplorer to display the countries most mentioned in the news or the latest stories about a country/place. Using other extracted information like person names, we can also display the persons most mentioned in news about a country/place, or the countries mentioned in stories about a defined person.

We would like to adapt our tool, in order to compare its performance with other approaches (as presented at MUC 1998 or GeoCLEF 2006). Currently the tool does not recognise places like streets, airports, etc.

Our database currently contains country names and cities only. We plan to add administrative unit names (California, Sicily, Bali, Cornwall etc.), and sea, river, island and mountain names (Channel, Danube, Elba, Mont-Blanc), which is another challenge because some of them belong to more than one country. This would also help us avoid errors. Examples are Lombardia (Brazilian city and Italian region) or Montenegro (city in Brazil but mainly used to design the republic of ex Yugoslavia).

### 7. Acknowledgment

Many thanks to Peeter Päll for providing us with the KNAB database, and all the persons who contributed with their language-specific knowledge.

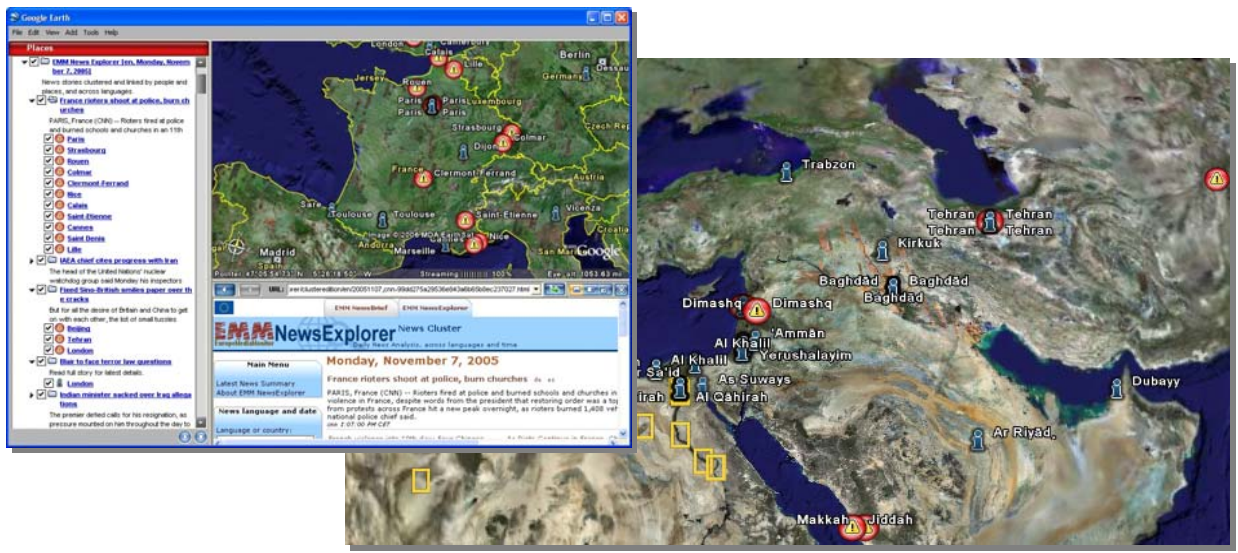

Figure 2: Geographical map browsing using Google Earth, the left image showing the various ways to browse the information, the right one showing how we can highlight the news about a certain region of the world.

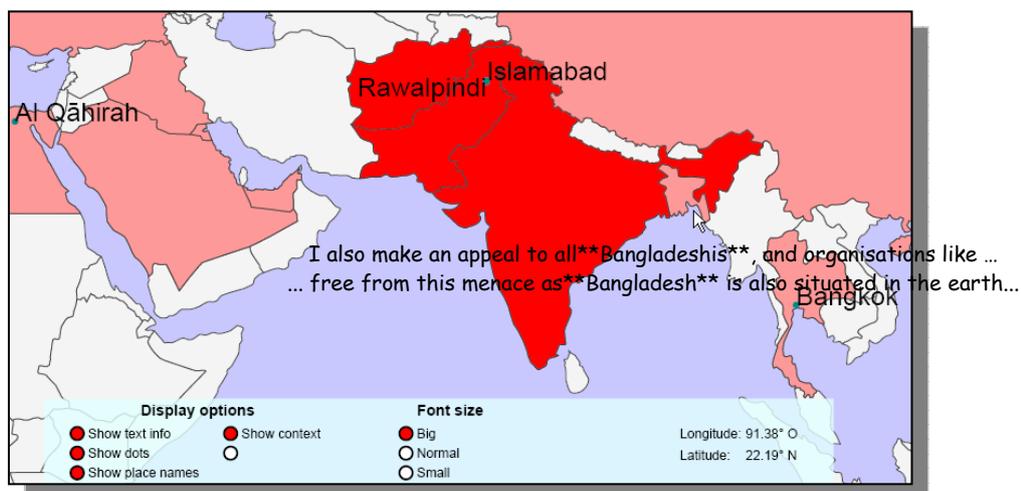

Figure 3: Geographical map using interactive SVG maps which can display the text passages where the place was found (here Bangladesh).